\documentclass[lettersize,journal]{IEEEtran}

\usepackage{amsmath,amsfonts}
\usepackage{algorithmic}
\usepackage{algorithm}
\usepackage{array}
\usepackage[caption=false,font=normalsize,labelfont=sf,textfont=sf]{subfig}
\usepackage{textcomp}
\usepackage{stfloats}
\usepackage{url}
\usepackage{verbatim}
\usepackage{graphicx}
\usepackage{multirow} 
\usepackage{multicol}
\usepackage[hidelinks]{hyperref}
\usepackage{cite}
\usepackage{tikz,xcolor}
\usepackage{bm}
\usepackage{fancyhdr}
\usepackage{booktabs}
\usepackage{lettrine}
\usepackage{verbatim}
\usepackage{threeparttable}
\usepackage{makecell}
\usepackage{xcolor}
\usepackage[T1]{fontenc}
\usepackage{cite}
\definecolor{lime}{HTML}{A6CE39}
\DeclareRobustCommand{\orcidicon}{
	\begin{tikzpicture}
		\draw[lime, fill=lime] (0,0)
		circle[radius=0.16]
		node[white]{{\fontfamily{qag}\selectfont \tiny \.{I}D}}; \draw[white, fill=white] (-0.0625,0.095) circle [radius=0.007];
	\end{tikzpicture}
	\hspace{-2mm}
}
\foreach \x in {A, ..., Z}{%
	\expandafter\xdef\csname orcid\x\endcsname{\noexpand\href{https://orcid.org/\csname orcidauthor\x\endcsname}{\noexpand\orcidicon}}
}

\begin{document}

\title{DualGenerator: Information Interaction-based Generative Network for Point Cloud Completion}

\author{Pengcheng Shi\orcidA{}, Haozhe Cheng\orcidB{}, Xu Han, Yiyang Zhou and Jihua Zhu\orcidC{}% <-this % stops a space
\thanks{Manuscript received: April 28,2023; Revised: Jun 26, 2023; Accepted: August 18, 2023. }
\thanks{This paper was recommended for publication by Editor Cesar Cadena upon evaluation of the Associate Editor and Reviewers’comments. }
\thanks{This work was supported in part by the Key Research and Development Program of Shaanxi Province under Grant 2021GY-025 and Grant 2021GXLH-Z-097 (Corresponding author: Jihua Zhu).}
\thanks{The authors are with School of Software Engineering, Xi'an Jiaotong University, Xi'an 710049, China (E-mail:spcbruea@stu.xjtu.edu.cn; zhujh@xjtu.edu.cn). Code and trained models will be available at https://github.com/spc121/DualGenerator. }%
\thanks{Digital Object Identifier (DOI): see top of this page.}

}

\markboth{IEEE ROBOTICS AND AUTOMATION LETTERS. PREPRINT VERSION. AUGUST 2023}%
{ Shi \MakeLowercase{\textit{et al.}}: DualGenerator: Information Interaction-based Generative Network for Point Cloud Completion}


\maketitle

\begin{abstract}
Point cloud completion estimates complete shapes from incomplete point clouds to obtain higher-quality point cloud data. Most existing methods only consider global object features, ignoring spatial and semantic information of adjacent points. They cannot distinguish structural information well between different object parts, and the robustness of models is poor. To tackle these challenges, we propose an information interaction-based generative network for point cloud completion ($\mathbf{DualGenerator}$). It contains an {upper} adversarial generation path and a {lower} variational generation path, which interact with each other and share weights. DualGenerator introduces a local refinement module in lower path, which captures general structures from partial inputs, and then refines shape details of the point cloud. It promotes completion in the unknown region and makes a distinction between different parts more obvious.{The upper path effectively provides support for completion by establishing a comprehensive distribution of complete point clouds, while the design of DGStyleGan enhances the robustness of network}. Qualitative and quantitative evaluations demonstrate that our method is superior on MVP and Completion3D datasets. The performance will not degrade significantly after adding noise interference or sparse sampling. 
\end{abstract}


\section{Introduction}
\IEEEPARstart{3}{D} point cloud is an intuitive representation of 3D scenes and objects. There is a lot of research in the fields of robotics, autonomous driving, 3D modeling, and so on \cite{9904825}. However, generated point clouds are usually sparse and incomplete due to sensor limitations, object occlusion, and noise. The importance of point cloud completion is thus reflected. It can estimate the complete shape from incomplete point clouds to obtain higher quality point clouds and provide data integrity support for downstream tasks such as classification and segmentation.

In recent years, a surging number of researchers have applied deep neural networks to point cloud completion. PCN \cite{yuan2018pcn} learns global features from incomplete point clouds above all and then completes point clouds from coarse to fine according to the feature information. This completion method, employing an encoder-decoder structure, has been further developed in subsequent research \cite{yu2021pointr} by numerous scholars. Nevertheless, these methods frequently neglect fine-grained local information, and they pay too much attention to the global characteristics of the object but ignore fine details \cite{wen2022pmp}. This leads to poor model performance.
\begin{figure}
	\centering
	\includegraphics[width=0.95\columnwidth]{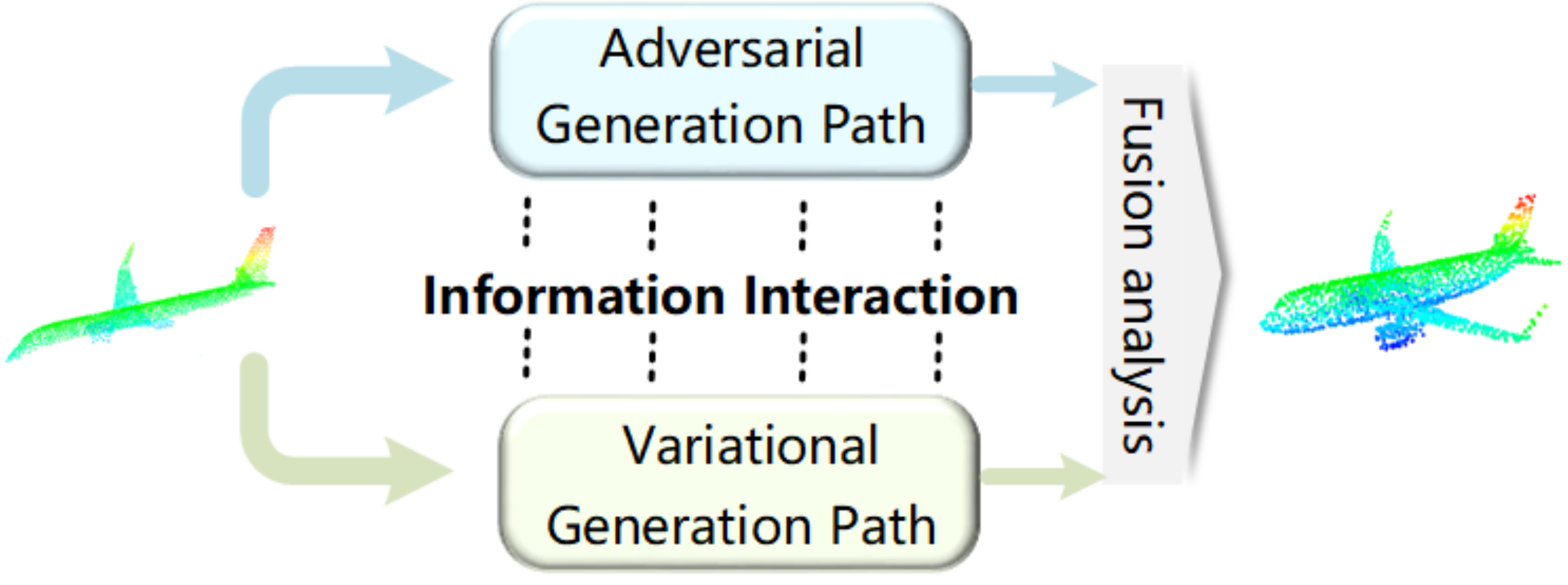} 
	\setlength{\belowcaptionskip}{-0.5cm}
	\caption{\textbf{Illustration of our DualGenerator}. It applies an interactive generative network with two pipelines to complete the shape of incomplete point clouds. Two pipelines, dubbed a variational generation path and an adversarial generation path, interact with each other and share weights. The completion point cloud is obtained by fusion analysis of the dual-path completion results.
	}
	\label{fig1}
	\vspace*{-0.6\baselineskip}
\end{figure}

To address these issues, we propose an Information Interaction-based Generative Network for Point Cloud Completion (entitled as \textbf{DualGenerator}), which contains an adversarial generation path \cite{an2015variational} and a variational generation path depicted in Fig. \ref{fig1}. It extracts features from the incomplete point cloud and its corresponding complete point cloud, respectively, where the encoder shares weights. The lower path distribution fits the upper path of this network. We design a local refinement module that can more faithfully restore the local details of the origin point cloud while preserving the spatial relations of adjacent points under the variational framework after sampling the feature distribution.

{Specifically}, we design DGStyleGAN to be more suitable for the task of point cloud completion \cite{karras2021alias}. {The generation method can decouple the latent space well. In addition, the optimization of convolutional layers realizes the translation and rotation invariance of point clouds, which enhances the robustness of our model.} What's more, the variational generation path combines the feature vectors of incomplete point cloud and coarse complete point cloud, as shown in Fig. \ref{fig2}. It enhances the structural relationship of point cloud objects by learning multi-scale local features. The model framework of VAE+DGStyleGAN can effectively handle the completion of 3D point clouds, and the generated point cloud structures from both models exhibit complementary characteristics. In the cause of reducing model sensitivity to noise points, we design a fusion analysis function that integrates the complete information of multi-scale relational decoder and style generator \cite{rahutomo2012semantic}. It significantly improves the robustness and applicability of this network.

The main contributions can be delivered as: 
\begin{itemize}
	
\item We propose a novel point cloud completion network (DualGenerator). It is an interactive generative framework with two pipelines, which can restore the unknown part and structure information of incomplete point clouds with high quality. 
\item We introduce a local refinement module in the network, which can capture structural features at a fine-grained level. Therefore, a distinction between different parts becomes more obvious. The semantic and spatial information is also well preserved between adjacent points. 
\item We design DGStyleGAN to improve the point cloud generation quality and fuse completion results of the generator with a multi-scale relational decoder, which promotes network robustness significantly.

\end{itemize}

\section{RELATED WORK}
\subsection{Point Cloud Completion}

Traditional point cloud completion methods are generally based on the prior information of natural structure. However, these methods can address some incomplete point clouds with low missing rates and obvious structural features merely, whose applicability is low. The development of deep learning has greatly promoted the ability and generalization of point cloud completion. Early point cloud completion methods attempt to transfer mature methods from 2D completion tasks to 3D point clouds by voxel localization \cite{9944850} and 3D convolution \cite{wu2019pointconv}, but the computational cost is high. The direct application of 3D coordinates has become the mainstream of point cloud tasks with the great success of PointNet \cite{qi2017pointnet} and PointNet++ \cite{qi2017pointnet++}. PCN \cite{yuan2018pcn} is first proposed for point cloud completion, but it only learns the common features of a certain type of object, which lacks the ability to reconstruct specific structures. Many scholars have made improvements to PCN subsequently. SK-PCN \cite{nie2020skeleton},  MSPCN \cite{zhang2020multi}, FinerPCN \cite{chang2021finerpcn} and other algorithms \cite{9312445} further improve the accuracy of point cloud completion by considering local information and reducing structural dependence. However, such methods are not effective for input with fuzzy structures and asymmetry.

{To solve this problem}, other point cloud completion methods are proposed. GRNet \cite{xie2020grnet} introduces 3D mesh as an intermediate representation to normalize the unordered point cloud, thus explicitly preserving the structure and background of a point cloud to capture the spatial relationship of a point cloud. However, some local structural details could not be recovered. PointLIE \cite{zhao2021pointlie} proposes a novel locally invertible embedding for point cloud adaptive sampling and recovery. {ShapeFormer \cite{yan2022shapeformer} can sample the distribution of results to generate plausible completions, with each completion exhibiting reasonable shape details while remaining faithful to the input.} SeedFormer \cite{zhou2022seedformer} proposes a novel network to improve the ability of detail preservation and recovery in point cloud completion. Unlike previous methods based on a global feature vector, it introduces a new shape representation, namely patch seeds, which captures not only general structures from partial inputs but also preserves regional information of local patterns. This provides a great idea for the design of our point cloud network.
\subsection{{Generative Model}}

Generative models attract attention with the deepening of research \cite{xie2021style} increasingly. Variational autoencoder structures, GAN-based methods and diffusion-based generative methods \cite{zhou20213d} are the most widely used ones at present. The model generated by GAN-based methods is relatively complete and has better anti-noise, but its trained network on 3D shapes generates point clouds with significant inhomogeneity, which destroys the integrity of predictions and jeopardizes reconstruction tasks \cite{valsesia2020learning}. Diffusion-based generative methods excel at generating high-quality data with structural consistency. They demonstrate proficiency in handling various shapes but face challenges with complexity and large-scale data, being sensitive to noise and incompleteness \cite{yang2022diffusion}. The good news is that variational autoencoders are well-suited for learning a well-structured latent space, which is more evenly distributed \cite{von2017advances}. VRCNet \cite{pan2021variational} proposes a variational relational point completion network recently, which designs the probabilistic modeling network and relational enhancement network. It can be used to recover complete point clouds from coarse to fine. However, its completion results are not uniform and lack local domain details.

\begin{figure*}
	\centering
	\includegraphics[width=0.95\textwidth]{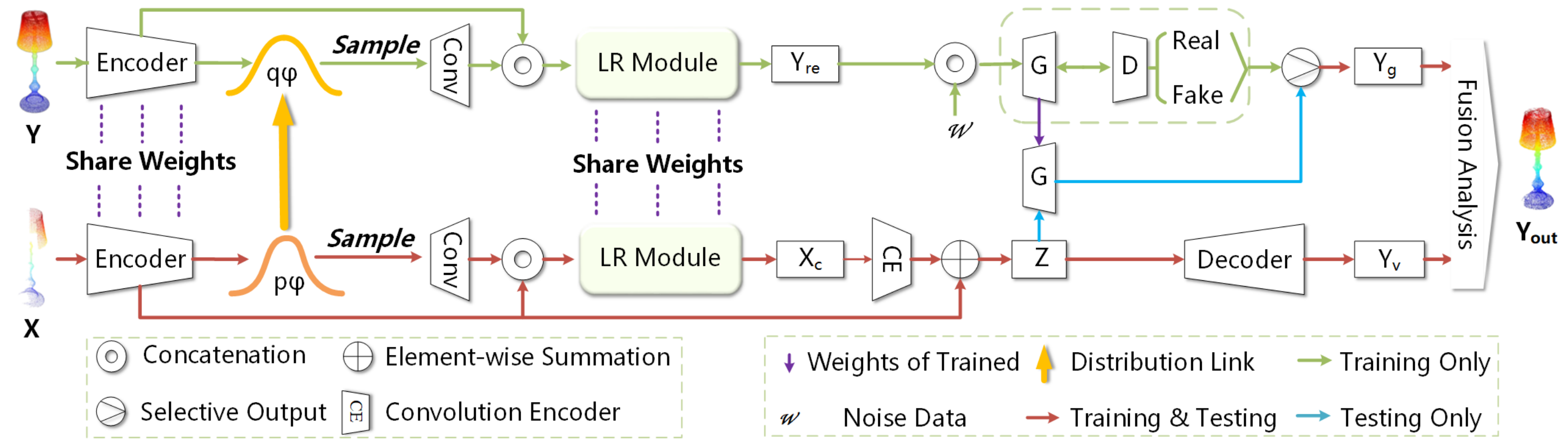} 
	\caption{\textbf{Structure of our DualGenerator}, which consists of an upper adversarial generation path and a lower variational generation path. The lower path obtains a latent space $\mathbf{Z}$, which decodes to get a completion point cloud $\mathbf{{Y}_{v}}$, based on the inferred distribution and encoder features. {During the training phase, $\mathbf{{Y}{re}}$ and $\textit{w}$ are used as inputs to the DGStyleGan model to learn the weights of the generator and discriminator.} It generates a new completion point cloud $\mathbf{{Y}_{g}}$ during testing. A fusion analysis function calculates the deviation degree of dual-path completion point clouds, and then fusions to get the result.}
	\label{fig2}
	\vspace{-.3cm}
\end{figure*}

\section{OUR METHOD}

The overall architecture of DualGenerator is shown in Fig. \ref{fig2}. We will introduce our method in detail as follows.
\vspace{-.2cm}

\subsection{{Overall Architecture}} DualGenerator defines the incomplete point cloud $\mathbf{X}$ as a partial observation for a 3D object, and a complete point cloud $\mathbf{Y}$ is sampled from the surfaces of an object. DualGenerator consists of an upper adversarial generation path and a lower variational generation path. {To be specific, the adversarial generation path provides support for completion by establishing a comprehensive distribution of complete point clouds.} The variational generation path can predict a completion point cloud $\mathbf{{Y}_{v}}$ containing local details and structural information combined with the point cloud features obtained by encoders. During testing, we take the latent space $\mathbf{Z}$ obtained in the lower path, which is uniform and well-structured, as input to the style generator, so as to obtain the completion point cloud $\mathbf{{Y}_{g}}$ with details of origin shape. A fusion analysis function is used to evaluate the deviation degree of $\mathbf{{Y}_{g}}$ and $\mathbf{{Y}_{v}}$ to pursue better generation quality and improve the robustness of completion point clouds. According to the deviation result of point clouds, we design different fusion methods to get the completion result.

Denote the input point cloud as $\boldsymbol{P}=\{\mathbf{p}_i|i=1,2,...,N\} \in \mathbb{R}^{N \times 3}$ where $N$ is the total number of points and $\mathbf{p}_i$ possesses the $(x,y,z)$ coordinates. The encoder extracts patch feature $\boldsymbol{{F}_p} \in \mathbb{R}^{N_p \times S_p\times D_p}$ and corresponding patch coordinate $\boldsymbol{{P}_p} \in \mathbb{R}^{N_p \times 3 \times D_p}$ from object point clouds by setting abstraction layers. {Here, $N_p$ represents the number of sampled points, $S_p$ and $D_p$ denote the dimensions of the features.} According to the dimension of input point clouds, the number of points obtained from each layer is controlled to decrease gradually. Moreover, the attention layer is designed to strengthen the connection between $\boldsymbol{{F}_p}$ and $\boldsymbol{{P}_p}$, taking advantage of spatial and geometric relationships between the neighbors of each node to enhance coding features well.

The encoder shares weights for incomplete point cloud $\mathbf{X}$ and corresponding complete point cloud $\mathbf{Y}$. We infer the distributions $\,q_{\phi}(\mathbf{z_g}|\mathbf{Y})$ and $p_{\psi}(\mathbf{z_g}|\mathbf{X})$ based on encoded information $\mathbf{z_g}$. To obtain significant and comprehensive features from incomplete point clouds, the variational generation path makes the conditional distribution $p_{\psi}(\mathbf{z_g}|\mathbf{X})$ fitting the potential distribution $\,q_{\phi}(\mathbf{z_g}|\mathbf{Y})$ during training (as shown in Fig. \ref{fig2}, the distribution link arrows indicate we hope distribution $p_{\psi}(\mathbf{z_g}|\mathbf{X})$ fitting $\,q_{\phi}(\mathbf{z_g}|\mathbf{Y})$ ). Therefore, $\,q_{\phi}(\mathbf{z_g}|\mathbf{Y})$ constitutes the prior distribution, $p_{\psi}(\mathbf{z_g}|\mathbf{X})$ is the posterior conditional distribution. We use KL divergence as the target function for distribution fitting:

\vspace{-.2cm}
\begin{equation}
	% \small{
		\begin{aligned}
			\mathcal{L}^{adv}_{KL} = &- \, \mathbf{KL}\big[q_{\phi}(\mathbf{z_g}|\mathbf{Y}) \, \big\| \,p(\mathbf{z_g})\big], \\
		\end{aligned}
		% }
\end{equation}
\vspace{-.2cm}
\begin{equation}
	% \small{
		\begin{aligned}
			\mathcal{L}^{var}_{KL} = &- \, \mathbf{KL}\big[p_{\psi}(\mathbf{z_g}|\mathbf{X}) \, \big\| \,q_{\phi}(\mathbf{z_g}|\mathbf{Y})\big], \\
		\end{aligned}
		% }
\end{equation}
\vspace{-.2cm}

\noindent where KL denotes KL divergence. $p(\mathbf{z_g})=\mathcal{N}(\mathbf{0},\mathbf{I})$ is a conditional prior predefined as Gaussian distribution. $\mathcal{L}^{adv}_{KL}$ and $\mathcal{L}^{var}_{KL}$ are used as the distribution fitting loss of adversarial generation path and variational generation path, respectively. It is worth mentioning that the ground truth in the adversarial generation path is only used for training, so the dual path architecture does not affect our reasoning efficiency.

{The point cloud distribution obtained through sampling is refined by a local refinement module, resulting in a coarse representation that encompasses the complete structure. By combining this representation with the features obtained from the encoder, the latent space $\mathbf{Z}$ is computed, which is uniform and well-structured.} To recover local shape details, we use a relational enhancement network (RENet)\cite{pan2021variational} as the decoding module. It uses the residual selection kernel module (R-PSK) as the building block. A hierarchical encoder-decoder architecture with edge retention pool (EP) and edge retention solution pool (EU) modules are adopted. It can effectively learn multi-scale structural relationships to generate high-resolution complete point clouds with predicted fine local details. The decoder is generated from coarse to fine to complete the point cloud. After obtaining coarse completions $\mathbf{{Y}^{c}_{v}}$,  RENet targets at enhancing structural relations to recover local shape details to obtain $\mathbf{{Y}_{v}}$ with higher completion accuracy. {Finally, the completion result $\mathbf{{Y}_{out}}$ is computed through a fusion analysis function.}

\begin{figure*}
	\centering
	\includegraphics[width=0.78\textwidth]{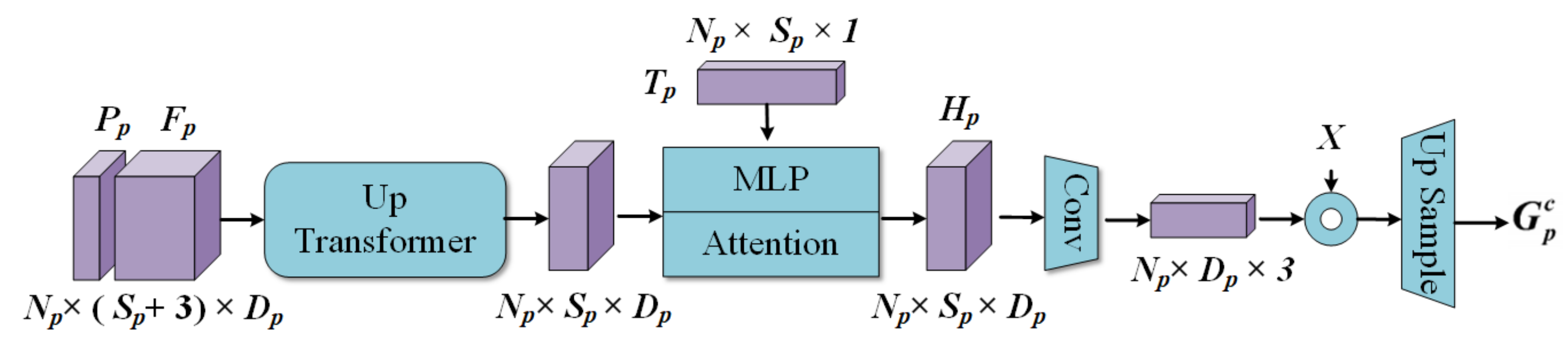} 
	\caption{\textbf{The local refinement module}, which can obtain details and structural characteristics of the point cloud with high accuracy. This module creates new points to cover the unknown part of an incomplete point cloud above all and then generates dense points in the local domain while emphasizing structural details and semantic information between points. In addition, the mathematical representation in this figure represents the shapes of different features.}
	\label{fig3}
	\vspace{-.3cm}
\end{figure*}

\begin{figure}
	\centering
	\includegraphics[width=0.35\textwidth]{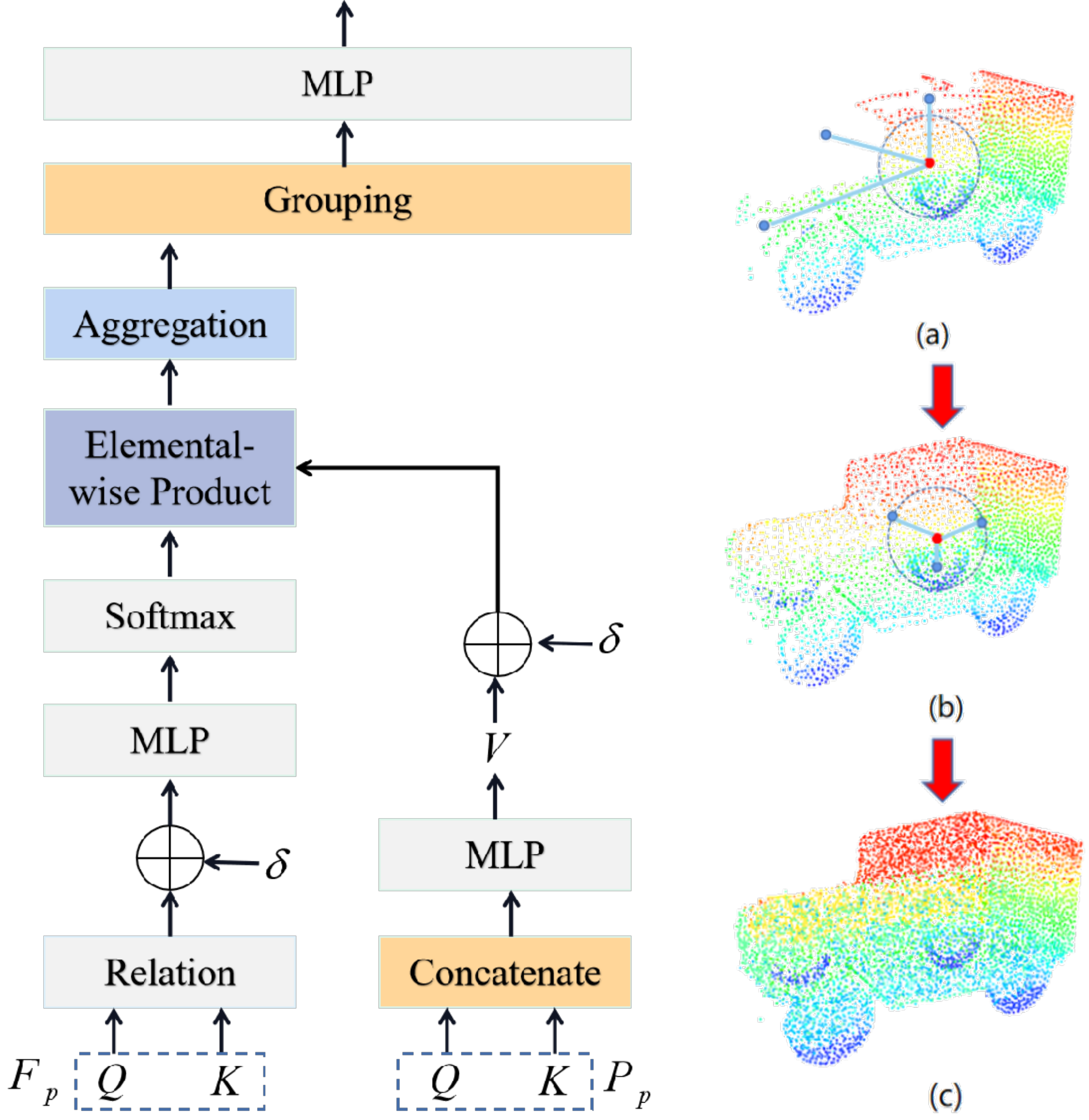} 
	\caption{\textbf{{Structural design and visual effects of Up Transformer.}}
		(a) Generate new points in the unknown parts of incomplete point clouds by Up Transformer. (b) Complete point clouds in the neighborhood of selected central points to generate denser results. (c) The coarse completion point cloud after local refinement.
	}
	\label{fig4}
	\vspace{-.6cm}
\end{figure}

\subsection{Local Refinement Module}

The encoder obtains patch coordinates $\boldsymbol{{P}_p}$ and patch features $\boldsymbol{{F}_p}$ of each layer, which contains semantic clues about a small area. Then {$\boldsymbol{{T}_p} \in \mathbb{R}^{N_p \times S_p\times 1}$} is obtained by sampling the fitted gaussian distribution. These three features are used as the input of a local refinement module (entitled as LR Module), as shown in Fig. \ref{fig3}.{We compute the local region features of adjacent points based on the obtained feature vectors. Subsequently, It gradually generates new points within the point cloud shape by employing an upsampled weighted averaging method.} After generating new points in the unknown part of incomplete point clouds, the fully connected layer of self-attention is applied to incorporate closely related local information into the generation combined with distribution sampling $\boldsymbol{{T}_{p}}$. It can be expressed as the following formula:

\vspace{-.2cm}
\begin{equation}
	% \small{0
		\boldsymbol{{H}_{p}} = {\alpha_{att}}({UpTrans}(\boldsymbol{{F}_p}, \boldsymbol{{P}_p}),\boldsymbol{{T}_p} ),
		% }
\end{equation}

\vspace*{0.4\baselineskip}
\noindent where $\alpha_{att}$ is the feature mapping function containing the self-attention mechanism to generate an attention vector. $\textit{UpTrans}$ represents a upsample transformer that generates new points around a target point as shown in Fig. \ref{fig4} (a).

{The design of Up Transformer is shown on the left in Fig. 4. Utilizing a dual-ended encoder design, we represent the $\boldsymbol{{P}_p}$ connection as the 'Value' and introduce positional encoding $\boldsymbol{\delta}$  . The element-wise product is computed by leveraging the feature representation of $\boldsymbol{{F}_p}$ . Through the process of aggregation, we generate novel representations of grouped patch points alongside their corresponding features.}

In addition, we construct a convolution layer to preserve fine features of the local region following the idea of local feature aggregation. For recovering the full shape structure from explicitly supervised partial inputs, combined with the input incomplete point cloud $\mathbf{X}$, an upsample sampler is used to recover a locally finely described point cloud $\boldsymbol{{G}^{c}_{p}}$. It effectively combines the spatial relationship between adjacent points, strengthens the discrimination between different parts, and faithfully reflects the local details of an object point cloud. Meanwhile, completion feature $\boldsymbol{{H}_{p}}$ contains the distribution performance of incomplete point cloud after fitting, which makes the generated $\boldsymbol{{G}^{c}_{p}}$ fit the structure shape of complete point clouds well:

\begin{equation}
	% \small{0
		\boldsymbol{{G}^{c}_{p}} = UpSample(\beta(\boldsymbol{{H}_p}), \mathbf{X}),
		% }
\end{equation}

\noindent where $\beta$ represents the convolution layer and $\mathbf{X}$ indicates a incomplete point cloud. $\textit{UpSample}$ means that up-sampling is completed by interpolation after extracting pointwise features for $\boldsymbol{H}_{p}$  as shown in Fig. \ref{fig4} (b).

\subsection{{Robust Reinforcement Structure}}

Although the variational generation path that consists of a local refinement module restores incomplete point clouds with high accuracy, the effect is not satisfactory to input noise. The adversarial generation path can correct the shape of a point cloud completed by the lower path, which greatly reduces the probability of generating deviation points and {improves the robustness of our method}. We design DGStyleGAN, which is specifically tailored for point cloud completion, and it offers several refinements compared to StyleGAN3. The key optimizations are outlined below:
\begin{itemize}
	 \item We replace 3 $\times$ 3 convolutions with 1 $\times$ 1, and increase the depth of feature mapping to enhance information acquisition. 
	 \item We preserve the input noise $\textit{w}$ and performe engineering sampling operations. 
	 \item We apply the low pass filter, which optimizes boundary and upsampling problems.
\end{itemize}

DGStyleGAN realizes the invariance of point cloud translation and rotation by changes in the convolution layer, which promotes generation quality. Moreover, the style generator is not limited to the structure of traditional "convolution + nonlinear layer + upsampling" but achieves excellent equivariance based on a scale-specific level of control.

{During the training phase, $\mathbf{{Y}_{re}}$ and $\textit{w}$ are used as inputs to the DGStyleGan model to learn the weights of the generator and discriminator. By utilizing the generated real point cloud and fake point cloud, the complete point cloud features are computed, resulting in $\mathbf{{Y}_{g}}$ that encompasses the comprehensive characteristics of the point cloud.} The latent space $\mathbf{Z}$ constructed by variational adversarial path can provide a well-distributed input to the style generator with shared weights during testing, and completion point cloud $\mathbf{{Y}_{g}}$ is obtained by the generator:

\begin{equation}
	\mathbf{{Y}_{g}}= \begin{cases} w_g\, Real +( 1 - {w}_g)\, Fake, & \text { \textit {Training}} \\ 
		Generator(\mathbf{Z} ), & \text { \textit {Testing}} \end{cases},
\end{equation}

\noindent where ${w}_{g}$ is the weighted parameter, $\mathbf{Z}$ represents latent space. $\textit{Real}$ and $\textit{Fake}$ represents real point clouds and virtual point clouds. $\textit{Generator}$ indicates generator in DGStyleGAN.

We design a fusion analysis function to estimate the deviation of generated point clouds in dual paths. Firstly, we use an L2 norm to constrain point clouds and then calculate their cosine distance. We adjust the threshold dynamically to correct the deviation points in generated point clouds according to the number of iterations and cosine distance. Then, we design learnable weights according to different similarities, which improves the accuracy and robustness of our model:

\begin{small}
	\begin{equation}
		% \small{0
			\mathbf{S} = \operatorname{dist}(\mathbf{{Y}_{g}}, \mathbf{{Y}_{v}})
			=1-\cos (\mathbf{{Y}_{g}}, \mathbf{{Y}_{v}})
			= 1- \frac{\mathbf{{Y}_{g}} \cdot \mathbf{{Y}_{v}}}{\|\mathbf{{Y}_{g}}\|\|\mathbf{{Y}_{v}}\|},
			% }
	\end{equation}
\end{small}

\begin{equation}\label{huber}
	\mathbf{{Y}_{out}}= \begin{cases} \mathbf{{Y}_{v}},&  \mathbf{0 \le S < S_{t}}  \\ 
		{w}_{s} \,\mathbf{{Y}_{g}}  + (1 -  {w}_{s}) \, \mathbf{{Y}_{v}}, & \mathbf{S} \ge \mathbf{S_{t}} \end{cases},
\end{equation}

\noindent where ${w}_{s}$ is the weighted parameter and $\mathbf{S}$ represents cosine distance.  $\mathbf{{S}_{t}}$ indicates the distance threshold that we set. In a specific experiment, we generally take $\mathbf{{S}_{t}}$ as 0.1, which is decayed by 0.9 every 10 epochs.

The fusion analysis function constrains deviation points while preserving the overall shape structure.  It makes the generated network more applicable in the face of irregular graphics and incomplete point clouds containing noise.

\subsection{Loss Function}

Our method is trained end-to-end, and the training loss consists of three parts. The details are as follows.

\begin{itemize}
\item The KL divergence loss of distribution fitting, including $\mathcal{L}^{adv}_{KL}$ of adversarial generation path and $\mathcal{L}^{var}_{KL}$ of variational generation path:
\end{itemize}

\begin{equation}
	% \small{0
		\mathcal{L}_{KL} =\lambda_{KL} (\mathcal{L}^{adv}_{KL} + \mathcal{L}^{var}_{KL}).
		% }
\end{equation}

\begin{itemize}
\item The loss of completion for generated point clouds. For the sake of training efficiency, the distance between two disordered point sets is measured, which is represented by chamfered distance $\mathcal{L}_{CD}$:
\end{itemize}

\begin{small}
	\begin{equation}
		\small{
			\mathcal{L}_{\mathbf{CD}}(\mathbf{P}, \mathbf{Q}) = \frac{1}{|\mathbf{P}|}\sum_{x\in\mathbf{P}}\underset{y\in\mathbf{Q}}{\min}\|x-y\|^2 + \frac{1}{|\mathbf{Q}|}\sum_{y\in\mathbf{Q}}\underset{x\in\mathbf{P}}{\min}\|x-y\|^2, 
		}
		\nonumber
	\end{equation}
\end{small}

\begin{small}
	\begin{equation}
		\mathcal{L}_{\mathbf{CD}} =\lambda_{c}\mathcal{L}_{\mathbf{CD}}(\mathbf{{Y}^{c}_{v}}, \mathbf{Y})
		+\lambda_{f}\mathcal{L}_{\mathbf{CD}}(\mathbf{{Y}_{v}}, \mathbf{Y})
		+\lambda_{o}\mathcal{L}_{\mathbf{CD}}(\mathbf{{Y}_{out}}, \mathbf{Y}),
		\label{eq:cd_loss}
	\end{equation}
\end{small}

\noindent where $x$ and $y$ denote points that belong to two point clouds $\mathbf{P}$ and $\mathbf{Q}$, respectively. $\lambda_c$, $\lambda_f$, and $\lambda_o$ are trade-off coefficients.

\begin{itemize}
\item We use the square root of one-sided chamfer distance $\mathcal{L}_{P}$ to represent partial point cloud matching loss of the local refinement module:
\end{itemize}

\begin{equation}
	\begin{split}
		\mathcal{L}_{{P}} &=\lambda_{p}\mathcal{L}_{\mathrm{CD}^{\prime}}\left(\mathbf{{Y}_{out}}\rightarrow \mathbf{Y}\right)\\ &=  \lambda_{p}\sum_{\mathbf{p}_{i}^{1} \in \mathbf{{Y}_{out}}} \min _{\mathbf{p}_{i}^{2} \in \mathbf{Y}}\left\|\mathbf{p}_{i}^{1}-\mathbf{p}_{i}^{2}\right\|,
	\end{split}
\end{equation}

\noindent where $\mathbf{p}_{i}^{1}$ and $\mathbf{p}_{i}^{2}$ represent each point in different point clouds, $\lambda_{p}$ is the weighted parameter.

It is worth mentioning that generator loss and discriminator loss of DGStyleGAN are trained as branches separately. The generated point cloud $\mathbf{{Y}_{g}}$ is detached without backpropagation. The total loss function can be expressed as:

\begin{equation}
	% \small{0
		{L} =\mathcal{L}_{KL} + \mathcal{L}_{CD} + \mathcal{L}_{P}.
		% }
\end{equation}

\begin{table}[]
	\centering
	\vspace{0.2em}
	\caption{Point cloud completion on Completion3D dataset in terms of per-point L2 Chamfer distance $10^4$(lower is better).}
	\huge
	\renewcommand\arraystretch{1.5}
		\resizebox{0.99\columnwidth}{!}{
	\begin{tabular}{c|c|cccccccc}
		\hline
		Methods    & Average       & \multicolumn{1}{l}{Plane} & \multicolumn{1}{l}{Cabinet} & \multicolumn{1}{l}{Car} & \multicolumn{1}{l}{Chair} & \multicolumn{1}{l}{Lamp} & \multicolumn{1}{l}{Couch} & \multicolumn{1}{l}{Table} & Watercraft    \\ \hline
		PCN \cite{yuan2018pcn}        & 18.22         & 9.79                      & 22.70                       & 12.43                   & 25.14                     & 22.72                    & 20.26                     & 20.27                     & 11.73         \\
		TopNet \cite{tchapmi2019topnet}     & 14.25         & 7.32                      & 18.77                       & 12.88                   & 19.82                     & 14.60                    & 16.29                     & 14.89                     & 8.82          \\
		ECG \cite{pan2020ecg}        & 9.20          & 3.98                      & 14.23                       & 8.42                    & 10.59                     & 9.85                     & 12.69                     & 9.08                      & 5.80          \\
		PoinTr \cite{yu2021pointr}     & 8.26          & 3.68                      & 13.24                       & 8.15                    & 10.47                     & 9.68                     & 12.43                     & 8.06                      & 5.77          \\
		VRCNet \cite{pan2021variational}     & 8.12          & 3.94                      & 13.46                       & 6.72                    & 10.35                     & 9.87                     & 12.48                     & 7.73                      & 6.14          \\
		SeedFormer \cite{zhou2022seedformer} & 8.21          & 3.77                      & \textbf{11.92}              & 7.46                    & 9.85                      & 9.13                     & 11.74                     & 7.65                      & 5.55          \\
		PDR \cite{2021A}        & 8.07          & 3.56                      & 12.73                       & 7.02                    & 9.14                      & 8.29                     & \textbf{11.18}            & 7.13                      & 5.89          \\ \hline
		Ours       & \textbf{8.02} & \textbf{3.26}             & \underline{12.18}                       & \textbf{6.51}           & \textbf{9.02}             & \textbf{8.15}            & \underline{11.66}                    & \textbf{7.11}             & \textbf{5.49} \\ \hline
	\end{tabular}
}
\end{table}

% \usepackage{multirow}
\begin{table}[]
	\centering
	\vspace{0.2em}
	\caption{{Point cloud completion results on MVP and Completion3D datasets at the resolution of 2048 points. CD loss is multiplied by $10^4$. EMD loss is multiplied by $10^2$.} }
	\renewcommand\arraystretch{1.3}
	\resizebox{0.99\columnwidth}{!}{
		\begin{tabular}{c|ccc|ccc}
			\Xhline{0.5pt}
			\multirow{2}{*}{Methods} & \multicolumn{3}{c|}{MVP}                       & \multicolumn{3}{c}{Completion3D}               \\ \cline{2-7} 
			& CD$\downarrow$            & EMD $\downarrow$          & F1  $\uparrow$            & CD$\downarrow$            & EMD $\downarrow$          & F1  $\uparrow$            \\ \hline
			PCN \cite{yuan2018pcn}             & 9.78          & 1.95          & 0.322          & 18.22         & 5.36          & 0.216          \\
			TopNet \cite{tchapmi2019topnet}           & 10.19         & 2.44          & 0.305          & 14.25         & 3.96          & 0.237          \\
			ECG \cite{pan2020ecg}               & 7.06          & 2.36          & 0.443          & 9.20          & 3.52          & 0.317          \\
			PointTr \cite{yu2021pointr}          & 5.92          & 2.14          & 0.475          & 8.26          & 2.87          & 0.320          \\
			VRCNet~\cite{pan2021variational}            & 5.98          & 2.31          & 0.495          & 8.12          & 3.19          & 0.409          \\
			SeedFormer \cite{zhou2022seedformer}       & 5.85          & 1.86          & 0.488          & 8.21          & 2.66          & 0.348          \\
			PDR \cite{2021A}              & 5.72          & \textbf{1.47} & 0.498          & 8.07          & \textbf{2.09} & 0.442          \\ \hline
			Ours                     & \textbf{5.68} & \underline{1.53}         & \textbf{0.499} & \textbf{8.02} & \underline{2.13}          & \textbf{0.460} \\ \hline
		\end{tabular}
	}
\end{table}

\begin{table}
	\centering
	\vspace{0.2em}
	\large
	\caption{Shape completion results (CD loss multiplied by $10^4$) with various resolutions on MVP dataset.}
	\renewcommand\arraystretch{1.2}
	\resizebox{0.95\columnwidth}{!}{
		\begin{tabular}{l | cc | cc | cc}
			\Xhline{0.5pt}
			\multirow{2}{*}{\# Points}  & \multicolumn{2}{c|}{4096} & \multicolumn{2}{c|}{8192} & \multicolumn{2}{c}{16384} \\
			\cline{2-7}
			& \scriptsize CD $\downarrow$ & \scriptsize F1 $\uparrow$ & \scriptsize CD $\downarrow$ & \scriptsize F1 $\uparrow$ & \scriptsize CD $\downarrow$ & \scriptsize F1 $\uparrow$ \\
			\hline 
			PCN \cite{yuan2018pcn}               & 7.96       & 0.463       & 6.86 & 0.565           & 5.98        & 0.642       \\
			TopNet \cite{tchapmi2019topnet}            & 7.69       & 0.434       & 6.64 & 0.526           & 5.14        & 0.618       \\
			ECG \cite{pan2020ecg}              & 7.31       & 0.506       & 4.23 & 0.695           & 3.64        & 0.724      \\
			PoinTr \cite{yu2021pointr}          & 4.69       & 0.598       & 3.52 & 0.712           & 3.14        & 0.766       \\
			VRCNet \cite{pan2021variational}           & 4.82       & 0.629       & 3.68 & 0.720           & 3.21        & 0.783       \\
			SeedFormer \cite{zhou2022seedformer}     & 4.61       & 0.605       & 3.59 & 0.710           & 3.17        & 0.773       \\   
			PDR \cite{2021A}      & 4.42       & 0.638       & 3.45 & 0.739           & 2.91        & 0.801       \\
			\hline
			Ours  & \textbf{4.29} & \textbf{0.643} & \textbf{3.38} & \textbf{0.747} & \textbf{2.81} & \textbf{0.809} \\
			\Xhline{0.5pt}
		\end{tabular}
	}
	\vspace{-1mm}
	\label{tab:plain}
\end{table}

\section{EXPERIMENTS}

\subsection{Datasets}

We first evaluate DualGenerator on general completion benchmark MVP, MVP-40~\cite{pan2021variational,pan2021mvp} and Completion3D~\cite{8953650}. Completion3D is a large scaled 3D object dataset derived from the ShapeNet dataset. It concerns the completion task of sparse point cloud completion, where it generates only one partial view for each complete 3D object in ShapeNet dataset, and samples 2,048 points from the mesh surface for both the complete and partial shapes. MVP is computed based on 16 categories of CAD models. Twenty-six scanning angles are set for each model. It provides ground truth point clouds with different resolutions, including 2048, 4096, 8192, and 16384 points. {The MVP-40 dataset consists of 41600 training samples and 64168 testing samples from 40 categories in ModelNet40.   Its partial point clouds are sampled from complete point clouds with a pre-defined missing ratio, i.e., 50\% and 12.5\% missing.}  {In addition, we construct a dataset MVP-noise and a dataset MVP-missing to verify the robustness of our method.
} 
{We assess the reconstruction accuracy by computing the chamfer distance (CD) and earth mover's distance (EMD) between the predicted complete point cloud $\mathbf{{Y}_{out}}$ and the ground truth $\mathbf{Y}$. Additionally, we employ the F-score (F1) to evaluate the distance between object surfaces in the MVP dataset. This decision is based on the understanding that CD and EMD can be misleading due to their sensitivity to outliers. The F-score is defined as the harmonic average between precision and recall, and a higher F-score generally indicates better overall visual quality.}

\subsection{Implementation Details}

We train our network end-to-end using PyTorch implementation with V100 GPU. We use Adam optimizer with $\beta_1$ = 0.9 and $\beta_2$ = 0.999. The initial learning rate is 1$e^{-4}$ (decayed by 0.7 every 40 epochs). We set different values for weights of the loss function. In a specific experiment, $\lambda_{KL}$ is 20, $\lambda_{c}$ is 10, $\lambda_{f}$ is 1, $\lambda_{p}$ is 0.5, and $\lambda_{o}$ is an initial value of 0.01, which increases with the number of epochs.

\subsection{Comparison}
We exhaustively compare our model with seven baseline methods: PCN \cite{yuan2018pcn}, TopNet \cite{tchapmi2019topnet}, ECG \cite{pan2020ecg}, PoinTr \cite{yu2021pointr}, VRCNet \cite{pan2021variational}, SeedFormer \cite{zhou2022seedformer}, and PDR \cite{2021A}.

\vspace*{0.8\baselineskip}
\begin{itemize}
	\item \emph{Quantitative Comparison}
\end{itemize}

The quantitative comparison results of DualGenerator with other advanced point cloud completion methods on Completion3D dataset are shown in Table I, in which the DualGenerator achieves the best performance in terms of average chamfer distance across all categories.   The second bestpublished method leaderboard is PDR \cite{2021A}, which achieves 8.07 in terms of average CD. When considering per-category performance, DualGenerator achieves the best results in 6 out of 8 categories across all compared counterpart methods, which justifies the better generalization ability of DualGenerator across different shape categories.

{Table II shows the CD loss , EMD loss and  f-score on the Completion3D and MVP datasets with 2048 points. Table III presents the CD loss and f-score of various evaluation methods on the MVP dataset at different resolutions. The experimental results demonstrate that our method exhibits the overall best performance. Although it slightly lags behind the PDR \cite{2021A} method in terms of the EMD metric, our approach yields higher-quality completions for point clouds. Moreover, our method excels particularly at 4096 resolution, showcasing its advantage in completing incomplete point clouds. Compared to VRCNet, our completed point clouds exhibit superior performance in generating fine details. For sparse point cloud inputs, our method can systematically complete unknown regions and generate dense results.}

\vspace*{0.7\baselineskip}
\begin{itemize}
	\item \emph{Qualitative Comparison}
\end{itemize}

\begin{figure}[t]
	\centering
	\includegraphics[width=0.35\textwidth]{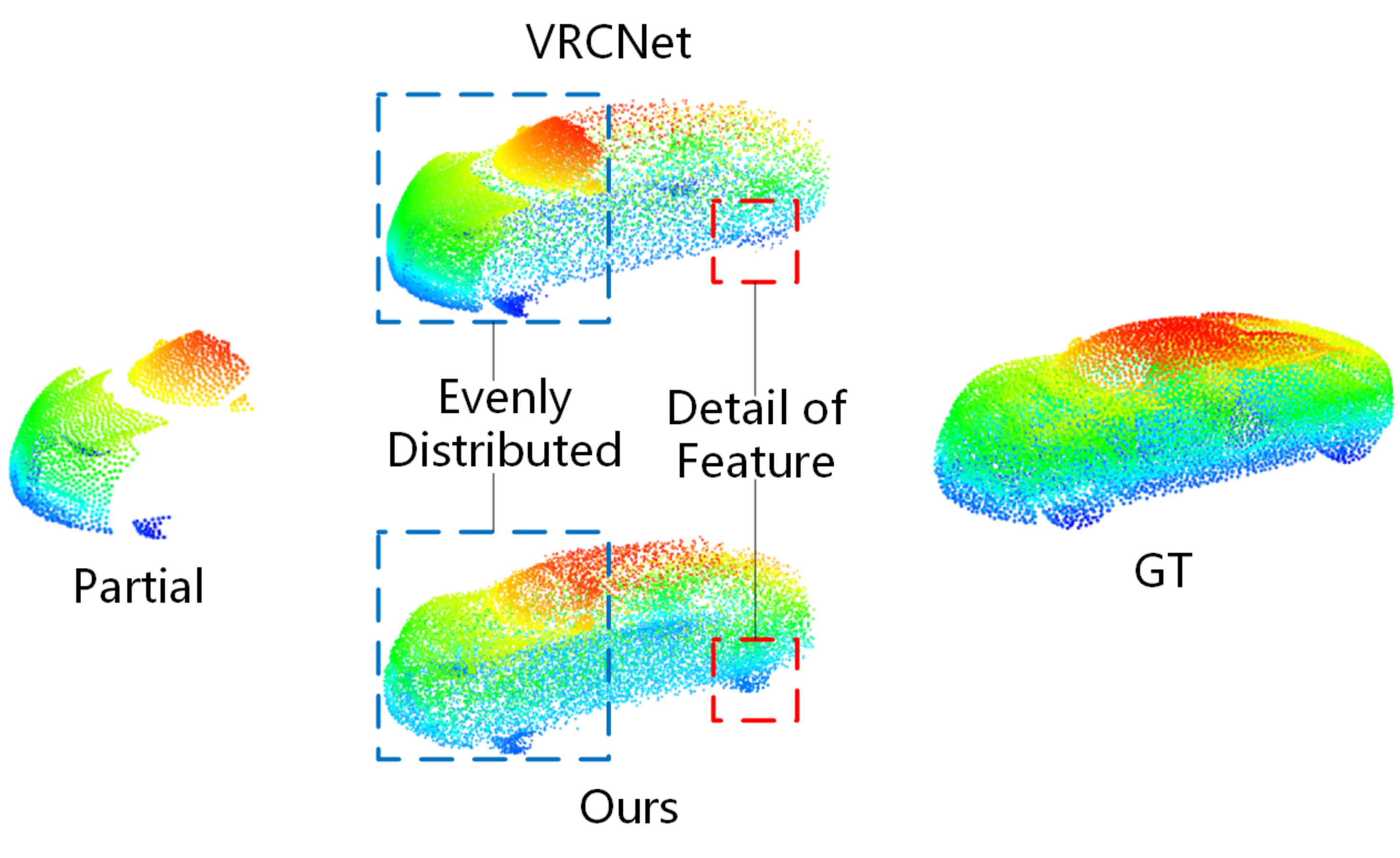} 
	\vspace{-.3cm}
	\caption{\textbf{Qualitative comparison.} Compare our method with baseline VRCNet on the model car of 16384 points. In terms of uniform distribution and detailed characteristics, our method has significant advantages on the completion point cloud.
	}
	\label{fig6}
	\vspace{-.3cm}
\end{figure}

\begin{figure*}
	\centering
	\includegraphics[width=0.78\textwidth]{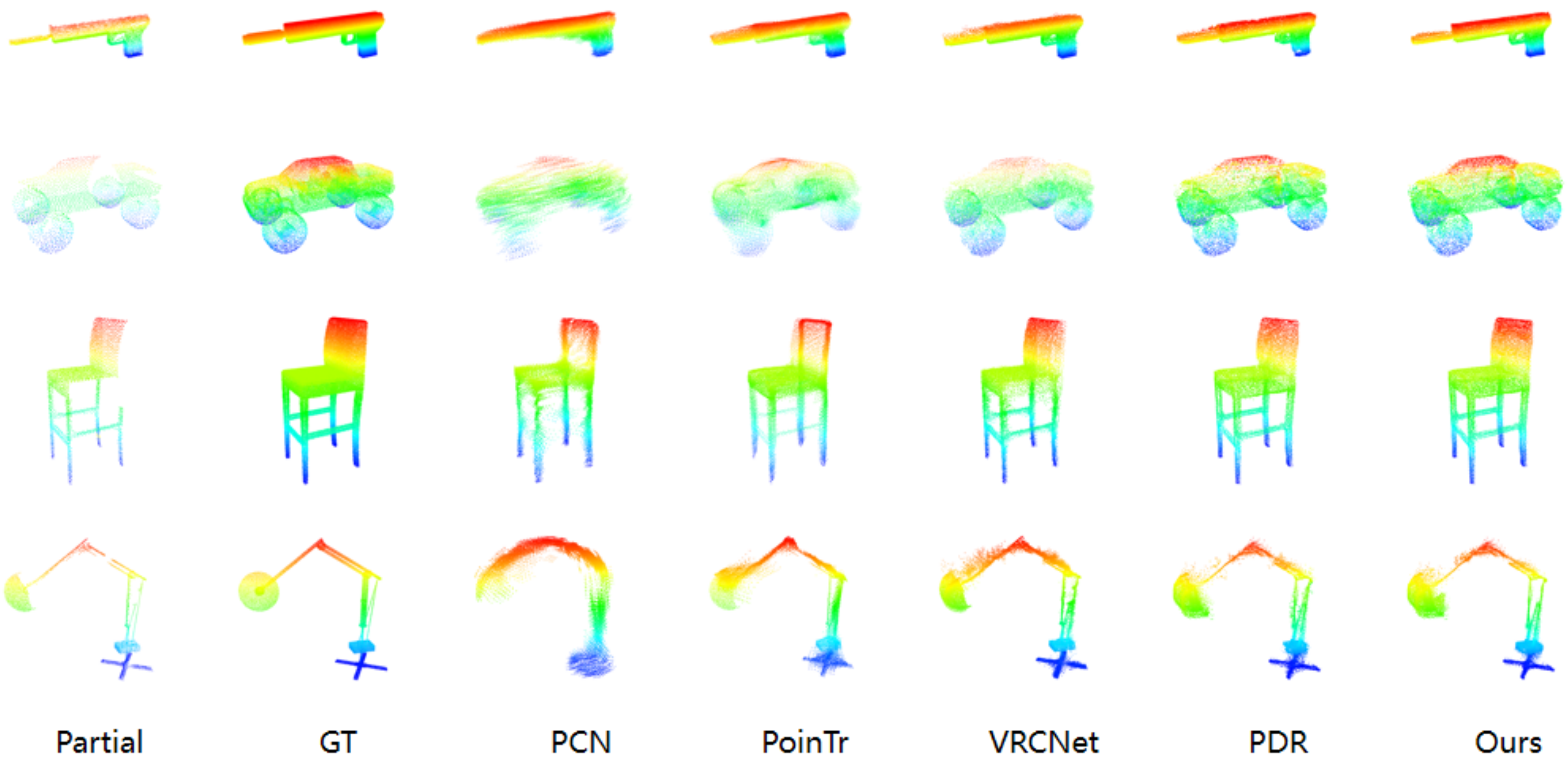}  
	\caption{\textbf{Visual comparison of completion results on benchmark datasets}, where our method can generate the best completion point clouds among all competing methods. Specifically, the point clouds generated by our method are relatively complete and have more structural information. In addition, the object parts are easier to be distinguished.}
	\label{fig5}
	\vspace{-.1cm}
\end{figure*}

We compare our method with other baseline methods visually on benchmark datasets in Fig. \ref{fig5}. In general, our method can preserve geometric details and local characteristics of partial inputs significantly. First, the point cloud boundary generated by our method is clearer during point cloud completion, which is crucial in restoring model details such as the barrel structure of a pistol and the top area of a car. Besides, our method can achieve a more uniform completion effect. It can learn the shape structure of unknown parts of incomplete point clouds and evenly distribute generated points of the overall model, as shown in Fig. \ref{fig6}, which can be closer to the ground truth. The most noteworthy point is that our method shows unique superiority in filling in missing details. According to the results of lamp and car generation, an unknown part of the lamp and wheel can be restored to a certain extent after training. We are excellent at capturing detailed features compared with other methods to recover model symmetry.

\begin{table*}[h]
	\centering
	\vspace{-1em}
	\caption{{Point cloud completion results on MVP-noise, MVP-missing datasets and MVP40 at the resolution of 2048 points.}}
	\renewcommand\arraystretch{1.5}
	%\resizebox{\textwidth}{22mm}{
		\setlength{\tabcolsep}{2mm}{
			\resizebox{2\columnwidth}{!}{
				\begin{tabular}{c|cccccc|cccccc|cccccc}
					\hline
					\multirow{3}{*}{Methods} & \multicolumn{6}{c|}{MVP-noise} & \multicolumn{6}{c|}{MVP-missing}& \multicolumn{6}{c}{MVP40}                               \\ \cline{2-19} 
					& \multicolumn{3}{c}{35\% noise points}  & \multicolumn{3}{c|}{10\% noise points} & \multicolumn{3}{c}{75\% missing} & \multicolumn{3}{c|}{50\% missing}& \multicolumn{3}{c}{50\% missing}& \multicolumn{3}{c}{12.5\% missing} \\
				   & CD $\downarrow$   & EMD $\downarrow$            & F1 $\uparrow$                & CD $\downarrow$    & EMD $\downarrow$             & F1 $\uparrow$                & CD $\downarrow$   & EMD $\downarrow$           & F1 $\uparrow$            & CD $\downarrow$   & EMD $\downarrow$           & F1 $\uparrow$        & CD $\downarrow$   & EMD $\downarrow$           & F1 $\uparrow$            & CD $\downarrow$   & EMD $\downarrow$           & F1 $\uparrow$      \\ \hline
					PCN~\cite{yuan2018pcn}         &12.22   &3.72   & 0.293                       & 10.36     &2.76              & 0.310                 & 10.62    &3.24            & 0.333              & 10.11   &3.10      &0.329    &39.67 &6.37  &0.581 &32.56 &6.18  &0.619       \\
					TopNet~\cite{tchapmi2019topnet}      & 12.77  &4.77    & 0.288               & 10.61    &3.21               & 0.302                 & 10.96    &4.06            & 0.291             & 10.66    &3.88       & 0.296  & 48.52 &8.75  & 0.506  & 40.12  &9.08  & 0.542             \\
					ECG~\cite{pan2020ecg}         & 9.18   &4.25    & 0.406                    & 7.47     &2.97               & 0.431                 & 7.47      &4.17          & 0.429              & 7.18   &4.02       &0.441       &34.06 &16.19  &0.671  &40.00 &6.98  &0.597        \\
					PoinTr~\cite{yu2021pointr}      & 8.06   &4.06    & 0.427                       & 6.34      &2.70              & 0.458                & 7.17    &3.82            & 0.443              & 6.88   &3.80             & 0.459              & 34.64 &15.78   & 0.613   & 26.03 &5.97   & 0.678  \\
					VRCNet~\cite{pan2021variational}      & 7.82  &4.33     & 0.447                      & 6.14  &3.08                  & 0.479                 & 6.74      &3.67          & 0.480              & 6.33     &3.51           & 0.486        & 25.70  &18.40    & 0.736     & 14.20  &5.90    & 0.807          \\
					SeedFormer~\cite{zhou2022seedformer}  & 8.03   &3.51    & 0.441                        & 6.27    &2.66                & 0.462                 & 6.77    &3.02            & 0.475              & 6.39       &2.67         & 0.481        &26.94 &9.74    & 0.706    & 13.56 &4.34    & 0.811         \\
					PDR~\cite{2021A} & 7.96  &2.24     & 0.447                      & 6.26         &1.65           & 0.463                 & 6.75     &2.66           & 0.482              & 6.34    &\textbf{2.32}           & 0.489               & 27.20 &2.68  & 0.739  & \textbf{12.70} &\textbf{1.39}  & 0.827 \\
					%					PMPNet++    & 6.97      & 0.461             & 6.31            & 0.472             & \textbf{5.71}          & 0.486                & 5.99               & 0.490              & 5.75               & \textbf{0.496}              
					\hline
					Ours       & \textbf{6.93} &\textbf{1.96}  & \textbf{0.463}    & \textbf{5.74}   &\textbf{1.60}  & \textbf{0.489}    & \textbf{5.97}  &\textbf{2.42}   & \textbf{0.492}    & \textbf{5.77}  &\underline{2.40}  & \textbf{0.494}  & \textbf{25.14} &\textbf{2.33}  & \textbf{0.774} &\underline{13.09} &\underline{1.67}  & \textbf{0.831}   \\ \hline
				\end{tabular}
		}}
		\vspace{-1mm}
		\vspace{-2mm}
	\end{table*}

\begin{figure}[t]
	\centering
	\includegraphics[width=0.48\textwidth]{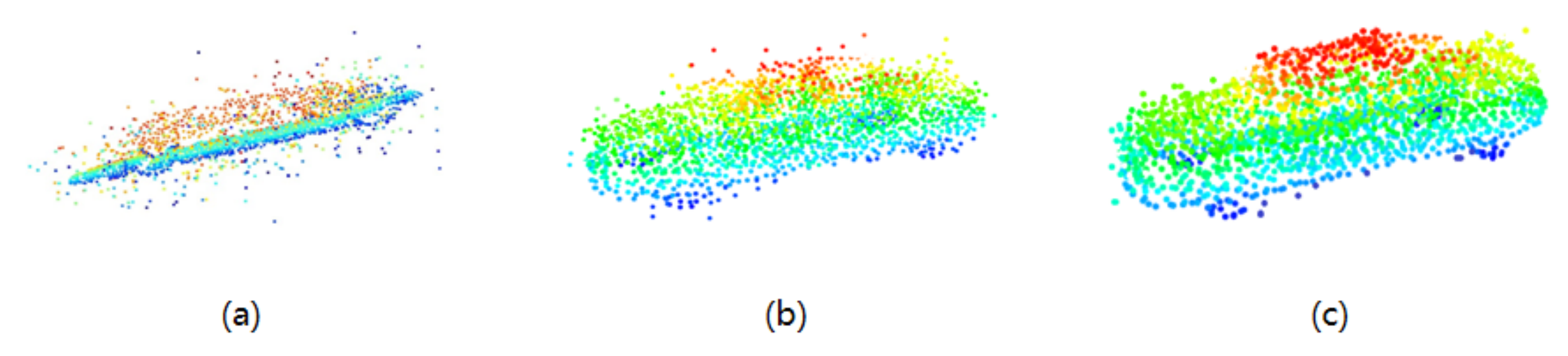} 
	\caption{\textbf{Coarse and fine point cloud completion results on MVP-noise dataset (2048 points).}
		(a) Incomplete point cloud of car containing 35$\%$ noise, which only includes the chassis shape of a car.
		(b) The coarse completion result of our method.
		(c) The fine completion result of our method.}
	\label{fig7}
	\vspace{-.3cm}
\end{figure}

\begin{table}[h]
	% \centering
	\vspace{2mm}
	\caption{Ablation studies (2,048 points) for proposed network modules on MVP dataset.}
	\renewcommand\arraystretch{1.1}
	\setlength{\tabcolsep}{2.0mm}{
		\begin{center}
			\footnotesize{
				\begin{tabular}{ccc|ccc}
					\Xhline{0.5pt}
					% \hline 
					% \scriptsize Point & \scriptsize Dual-pipeline & \scriptsize Kernel & \multirow{2}{*}{\scriptsize CD} & \multirow{2}{*}{\scriptsize F1} \\
					% \scriptsize Self-Attention & \scriptsize Architecture & \scriptsize Selection & & \\
					% \tabincell{c}{\footnotesize Point \\ \footnotesize Self-Attention} & \tabincell{c}{\footnotesize Dual-path \\ \footnotesize Architecture} & 
					% \tabincell{c}{\footnotesize Kernel \\ \footnotesize Selection} & \footnotesize CD & \footnotesize F1 \\
					\makecell{LR Module} & \makecell{DGStyleGAN} & \makecell{ $\mathcal{L}_{P}$ Loss} & 	\makecell{CD $\downarrow$} & 	\makecell{EMD $\downarrow$}& 	\makecell{F1 $\uparrow$}  \\
					\hline
					& \checkmark& \checkmark &  \footnotesize 6.02&  \footnotesize 2.46 & \footnotesize 0.491  \\
					\checkmark & & \checkmark&  \footnotesize 5.70 &  \footnotesize 1.97& \footnotesize 0.497  \\
					\checkmark & \checkmark & & \footnotesize 5.79 &  \footnotesize 2.23& \footnotesize 0.496 \\
					\checkmark & \checkmark & \checkmark & \footnotesize\textbf{ 5.68}& \footnotesize\textbf{1.53} & \footnotesize \textbf{0.499} \\
					% \hline
					\Xhline{0.5pt}
				\end{tabular}
			}
	\end{center}}
	\vspace{-5mm}
	% \vspace{-0.3cm}
	\label{tab:ablation_all}
\end{table}

\begin{table}[h]
	\setlength{\tabcolsep}{1.2mm}{
		\caption{Ablation studies (2,048 points) for GAN-based methods on MVP dataset, including only VG Path, GAN, StyleGAN3 and ours on MVP and MVP-Noise (35$\%$) datasets.}
		\renewcommand\arraystretch{1.1}
		\footnotesize
		\vspace*{-1.2\baselineskip}
		\begin{center}
			\begin{tabular}{c|ccc|ccc}
				\hline
				\multirow{2}{*}{GAN-based methods} & \multicolumn{3}{c|}{MVP}   & \multicolumn{3}{c}{MVP-noise(35\%)} \\
				& CD $\downarrow$   & EMD $\downarrow$         & F1 $\uparrow$              & CD $\downarrow$       & EMD $\downarrow$         & F1 $\uparrow$               \\ \hline
				only VG Path                & 5.70    & 2.17        & 0.497          & 7.67    & 4.23         & 0.449            \\
				GAN                           & 5.89     & 4.66     & 0.492          & 7.69   & 5.18          & 0.447            \\
				StyleGAN3                     & 5.72     & 1.97     & 0.496          & 7.07     & 3.88        & 0.456            \\ \hline
			Ours         & \textbf{5.68} & \textbf{1.53} & \textbf{0.499} & \textbf{6.93}  & \textbf{1.96}  & \textbf{0.463}   \\ \hline
			\end{tabular}
		\end{center}
	}
	\vspace{0.5mm}
	\centering('VG Path' : Variational Generation Path)
	\vspace{-5mm}
\end{table}

\subsection{Robustness Analysis}

We introduce additional datasets, MVP-noise and MVP-missing, by randomly generating noise points or removing several points from the MVP dataset. These datasets are used to verify the network's robustness. We compare our model with baseline methods under different proportions of missing or noise interference. {Additionally, we evaluate the performance of different models under MVP-40 with 50\% missing and 12.5\% missing.} The experimental results are summarized in Table IV. Our method exhibits the ability to generate completion point clouds with minimal accuracy loss, demonstrating excellent robustness and generalization capabilities when dealing with sparse or noisy input conditions. Refer to Fig. \ref{fig7} for visual evidence.

\subsection{Ablation Studies}
We evaluate the design of this network of our method here. We consider the following ablation versions: LR Module, DGStyleGAN, and $\mathcal{L}_{P}$ Loss. Experiments are conducted on MVP dataset at the resolution of 2048 points, and results are shown in Table V. We can see that the final version of our method has the best performance overall. 

We calculate results of the network without a local refinement module and directly take the feature distribution with results obtained by the encoder as \textbf{Z} to decode. The performance of our method drops when there is no local refinement module to refine the characterization. This shows the importance of an LR module in point cloud completion. In addition, the main contribution of DGStyleGAN is to reduce the sensitivity of our model to input noise. Experiments show that the design of $\mathcal{L}_{P}$ measures the loss in partial matching, which can maintain local features of point clouds preferably.

Besides, we do comparative experiments on the selection of different GAN-based methods, including only VG Path, GAN, StyleGAN3, and our final on MVP and MVP-Noise (35$\%$) datasets. As shown in Table VI, the introduction of an adversarial generation path can improve the model performance slightly. The main purpose we using GAN-based methods is to improve the robustness of our model. DGStyleGan can effectively reduce the interference of noise points and greatly improve the generalization performance of our method by the fusion analysis function.

\section{CONCLUSIONS}

In this paper, we propose an information interaction-based generative network for point cloud completion, featuring both an adversarial generation path and a variational generation path.  What sets our approach apart from previous point cloud completion methods is its ability to effectively preserve spatial information between adjacent points and enhance the discrimination of various parts.  This is achieved through the incorporation of a local refinement module in the dual paths, enabling the capturing of general structures from partial inputs, followed by the precise refinement of shape details in the completed point clouds.  Furthermore, we introduce DGStyleGAN to elevate the generation quality even further.  The comprehensive deviation degree analysis of our dual-path completion results showcases the robustness of our method in handling input noise and missing points.  Experimental results on the MVP, MVP-40, and Completion3D datasets demonstrate clear improvements over advanced competitors, establishing the effectiveness and superiority of our proposed approach.

\addtolength{\textheight}{-12cm}   % This command serves to balance the column lengths

%%%%%%%%%%%%%%%%%%%%%%%%%%%%%%%%%%%%%%%%%%%%%%%%%%%%%%%%%%%%%%%%%%%%%%%%%%%%%%%%
\bibliographystyle{IEEEtran}
\bibliography{IEEEtran}{}

\end{document}